\documentclass[letterpaper, 10 pt, conference]{ieeeconf}  

\IEEEoverridecommandlockouts                              

\usepackage{graphicx}
\usepackage{amsmath}
\usepackage{amssymb}

\title{\LARGE \bf
Exploring Social Motion Latent Space and Human Awareness for Effective Robot Navigation in Crowded Environments
}

\author{Junaid Ahmed Ansari, Satyajit Tourani, Gourav Kumar and Brojeshwar Bhowmick\\
 TCS Research, Kolkata, India \\
 \small{\{junaidahmed.ansari, satyajit.tourani, gourav.k6, b.bhowmick\}@tcs.com} }

\begin{document}

\maketitle
\thispagestyle{empty}
\pagestyle{empty}

\begin{abstract}

This work proposes a novel approach to social robot navigation by learning to generate robot controls from a social motion latent space. By leveraging this social motion latent space, the proposed method achieves significant improvements in social navigation metrics such as success rate, navigation time, and trajectory length while producing smoother (less jerk and angular deviations) and more anticipatory trajectories. The superiority of the proposed method is demonstrated through comparison with baseline models in various scenarios. Additionally, the concept of humans' awareness towards the robot is introduced into the social robot navigation framework, showing that incorporating human awareness leads to shorter and smoother trajectories owing to humans' ability to positively interact with the robot. 

\end{abstract}

\section{INTRODUCTION}

Social robotics has made significant progress in recent years, particularly in social robot navigation, which is crucial for robots to coexist with humans. However, this presents challenges such as navigating crowded and dynamic environments, understanding social cues, and ensuring safety, comfort, and efficiency. Effective social robot navigation requires planning and executing complex tasks while adapting to changing surroundings and meeting user expectations.

To address the above challenges, widely-used reactive methods include ORCA (Optimal Reciprocal Collision Avoidance)\cite{orca} and Social Force (SF)\cite{sf}. ORCA~\cite{orca} computes optimal velocities for agents in a centralized system to avoid collisions while pursuing goals. SF~\cite{sf} models interactions as forces to calculate movements. Learning-based approaches use MDP and neural networks for crowd navigation. A recent DS-RNN network has been proposed~\cite{dsrnn}, capturing spatio-temporal interactions between the robot and agents. Methods involving Deep V-Learning~\cite{rgl, cri}, approximate MDP state values using neural networks. They inherit ORCA's limitations during initialization (e.g., robot freezing) and assume deterministic human dynamics, which may not hold in real-world scenarios.

We present a novel approach for incorporating human awareness into the robot's social navigation framework. Humans are typically aware of a robot's presence and navigation in shared spaces, yielding to the robot, resulting in more efficient navigation with smoother trajectories. Prior research has largely overlooked this important aspect, despite its potential for improving social attributes.

CRI~\cite{cri} has high success rate but non-smooth trajectories with significant jerk. RGL~\cite{rgl} improves smoothness using a graph-based model. DSRNN~\cite{dsrnn} and CrowdNav++~\cite{intentaware} have smoother trajectories but longer navigation times.

In this paper we address the aforementioned challenges by making the following contributions:
\begin{enumerate}
    \item Our novel approach to social robot navigation incorporates the generation of actions from a social motion latent space of a trajectory forecaster. By leveraging this social motion latent space, we demonstrate significant enhancements in critical social navigation metrics, such as Success rate, Navigation time, Trajectory length. Our approach also results in trajectories that exhibit less jerk and greater anticipation. This helps us obtain smooth and social trajectories with a balance of faster navigation and shorter routes to goal.
    \item We compare our method to baseline models and illustrate its superiority through the following observations:
    \begin{enumerate}
    \item Despite not being trained using continuous actions, our method achieves state-of-the-art performance across various metrics compared to models such as Crowdnav++~\cite{intentaware} and DSRNN~\cite{dsrnn}, which are trained on continuous actions. We attribute this to our decision of using social motion latent space which is continuous in nature.
    \item Our approach outperforms DSRNN~\cite{dsrnn} and CrowdNav++~\cite{intentaware} across various metrics even when the robot's maximum speed is reduced (trained on a single max-speed(1m/s)).
    \item Despite not including a future collision check in the reward structure, unlike Crowdnav++~\cite{intentaware}, our approach is able to produce smoother robot trajectories with social behavior (respecting personal space by keeping sufficient distance from humans) as an emergent property.
    \end{enumerate}
    \item We introduce the concept of human awareness into the social robot navigation framework for crowded and dynamic environments. Our research demonstrates that humans tend to give way to robots when they are aware of the robot's presence, resulting in higher success rates, faster navigation, and shorter trajectories. Additionally, incorporating human awareness leads to better social behavior and smoother trajectories. 
    \end{enumerate}

\section{Related Work}

\subsection{Reaction-based methods}

ORCA~\cite{orca} and Socialforce~\cite{sf} are popular classical methods for robot navigation in dynamic crowds. ORCA~\cite{orca} models other agents as velocity obstacles to compute collision-free optimal velocities. However, it needs precise hyperparameter tuning and assumes all agents follow the reciprocal rule. SF~\cite{sf} uses attractive and repulsive forces to model interactions, considering social forces for optimal velocity computation.

\subsection{Learning-based methods}

CrowdNav~\cite{cri} and RGL~\cite{rgl} both use Deep V-Learning to model interactions between humans and robots for navigation purposes. While RGL~\cite{rgl} uses a graph-based attention mechanism for complex environments, both methods incorporate ORCA~\cite{orca} as an initialization step through imitation learning. This leads to it exhbiting the same limitations as ORCA.  DSRNN~\cite{dsrnn} overcomes the limitations of Deep V-Learning by utilizing a spatio-temporal graph to capture interactions between robots and humans, and requires no prior knowledge of agent dynamics or expert policies. CrowdNav++~\cite{intentaware} utilizes an intention-aware policy that considers interactions within a crowd and generates longer-sighted robot behaviors. However, all of these methods have either jerky or very long trajectories. 
\cite{samdrl} presents SA-CADRL, a deep reinforcement learning algorithm for socially aware multiagent collision avoidance. It outlines a strategy to shape normative behaviors in a two-agent system within the RL framework and subsequently extends the method to handle multiagent scenarios.
Frozone~\cite{frozone} presents a real-time algorithm that predicts pedestrian trajectories, identifies potential freezing situations, and constructs a conservative zone to avoid obtrusion. By combining this with a state-of-the-art DRL-based collision avoidance method, they achieve a significant reduction in freezing rates and improved pedestrian-friendly robot trajectories.

\subsection{Trajectory smoothness}
Achieving smooth trajectory planning is vital for efficient and safe robot navigation in complex environments. However, obstacles, mechanical constraints, and real-time considerations pose challenges. To address this, researchers have developed techniques like Model Predictive Control (MPC)\cite{mpc}, B-spline model with SCP method\cite{bspline}, and third-degree polynomial functions~\cite{jb}. These methods enable collision-free, length-optimal path planning with flexibility for higher-order smoothness while avoiding obstacles. Additionally, introducing a slight penalty for angular movements can enhance a robot's predictability in the presence of other agents~\cite{0386}.
\subsection{Social Awareness}
Socially-Aware Navigation~\cite{54} discusses challenges faced by robots in social environments. Detecting unattentive pedestrians, known as smartphone zombies, is addressed in~\cite{zombie}, an important safety concern. Gaze is a crucial aspect for collision avoidance~\cite{56},~\cite{57}. Human-robot cooperation in dense crowds is studied in ~\cite{59}, where the authors propose a solution to the Freezing Robot Problem (FRP) using an Interacting Gaussian Process (IGP) model of the crowd. ~\cite{60} reports a strong preference for humans to give way to the robot. Recent works~\cite{64},~\cite{65} have used humans' attention for robot navigation but do not address dynamic agents.

\begin{figure}

\center
\includegraphics[width=0.9\columnwidth]{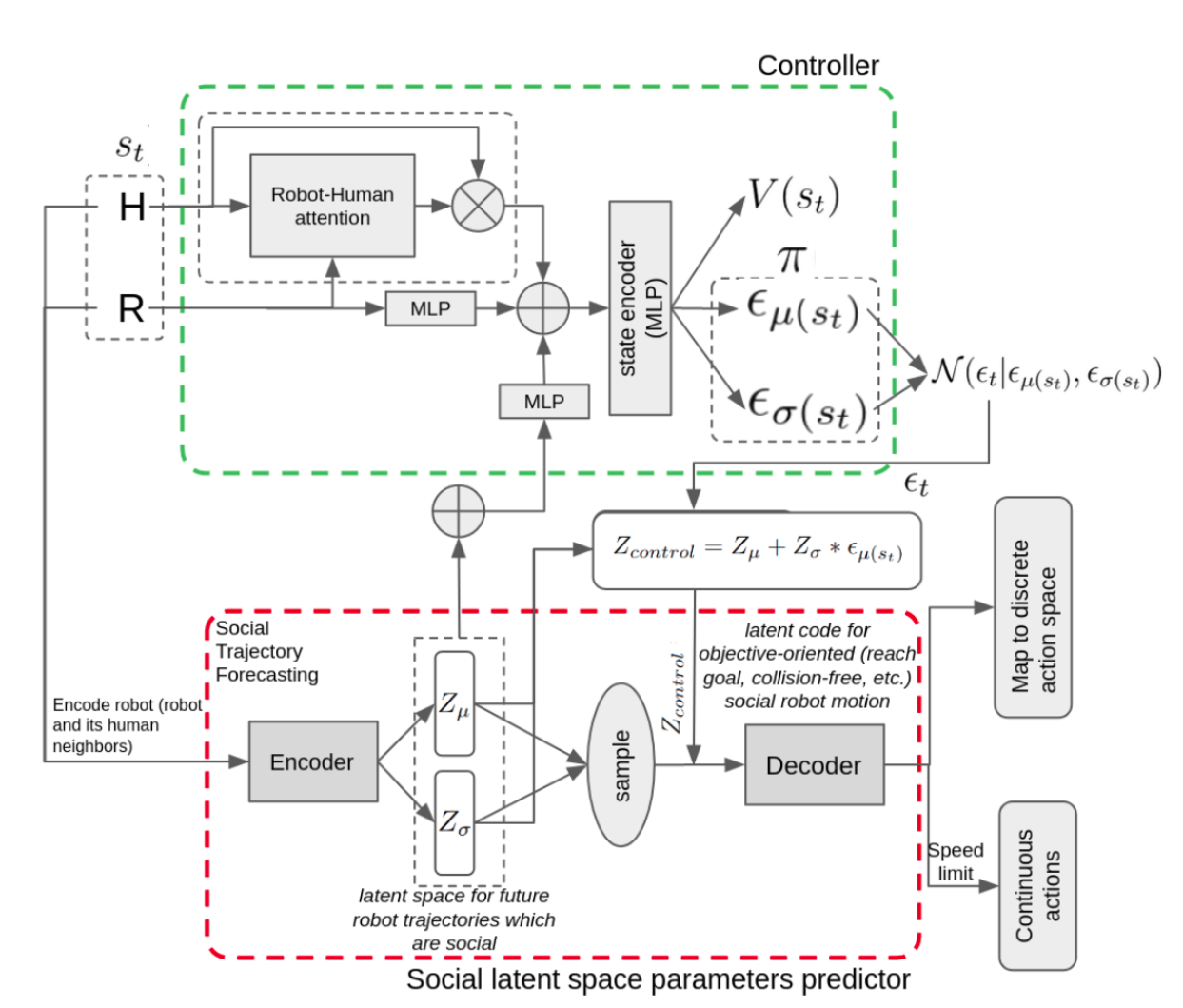}
  \caption{Model Architecture }
  \label{fig:2}
\end{figure}

\section{Methodology}

\subsection{Problem Formulation}

Consider a scenario with multiple dynamic humans and
a mobile robot which has to navigate from start pose to
goal pose. Let us assume that we have a multi-modal
trajectory forecaster (we use Social-VAE\cite{socvae}) which can
predict multiple future trajectories of a human which are
social in nature by selecting from a learnt social motion
latent space conditioned its own and neighbors’ past motion
information(tracking is assumed as we are operating in a simulation environment. However in reality methods like ~\cite{brojoda, boch}). This social latent space encodes how humans
move in reality (as seen in data during training). In this work,
we propose to exploit the social motion latent space of a
trajectory forecaster to generate goal-oriented, collision free
and human-like robot motion i.e. smooth, anticipatory and
respects personal-space of humans i.e. keeps a good distance
from humans. 

\subsection{Network architecture}
Our network as shown in Fig.~\ref{fig:2} is made up of two blocks: 1) a frozen social latent space parameter predictor(red) and 2) controller(green).

\subsubsection{Social latent space parameters predictor}

Our VAE~\cite{vae} based multi-modal social future trajectory predictor is trained on human crowd data. The encoder treats the robot as a human and predicts distribution parameters from a learned social motion latent space. The decoder converts sampled latent vectors to social, collision-free future motion. The motion is 'human-like' as it has learned from human crowd motion data. Although we use SocialVAE~\cite{socvae}, but any VAE~\cite{vae} based forecaster can also be used.

Formally, let $\bold{x}$ and $\bold{n}$ be the robot's  and its neighboring humans' past motion information. The encoder $\mathcal{E}$ takes $\bold{x}$ and $\bold{n}$ and produces $Z_\mu$ and $Z_\sigma$ (refer to work ~\cite{socvae} for more information). We can sample from the distribution $\mathcal{N}(Z_\mu, Z_\sigma)$ to produce a latent vector $Z$ which the decoder will use to predict a socially plausible motion vector for the next time instance relative to current position of the robot ($\Delta x$, $\Delta y$). 

\subsubsection{Controller}

We explained how to generate social motion using the social motion latent space of a trajectory forecaster. However, this lacks control over the generation process to enforce specific goals or norms. To create a social robot navigation policy as depicted in Fig.\ref{fig:2}, the green dashed box controller block learns a policy to shape the robot's behavior with different rewards and penalties, controlling the sampling process based on the social latent space.

Similar to \cite{motionvae}, the process of controlling the generation process is modeled as a Markov Decision Process (MDP) described by the tuple \textit{⟨$\mathcal{S}$, $\mathcal{A}$, $\mathcal{P}$, $\mathcal{R}$, $\gamma$, $\mathcal{S}$$_0$⟩}. Where $\mathcal{S}$ is the overall state of the environment which is derived from robot states, humans’ observable states and the social latent space parameters $(Z_\mu, Z_\sigma)$ corresponding to the robot.

\emph{State}: At any time instance $t$, the state $s_t$ is made up of robot's state (R) and all humans' observable states (H). let R be the robot state, comprising of position of the goal w.r.t robot $(px^r_g, py^r_g)$, relative velocity of goal w.r.t robot $(vx^r_g, by^r_g)$, sum of the radii of robot and human $(rad_r+rad_h)$, the angle between robot's velocity vector and goal vector w.r.t robot $theta_g$, and distance to goal $dg$. Similarly, let $h$ be human's state vector comprising of their position w.r.t. robot $(px^r_h, py^r_h)$, relative velocity w.r.t robot $(vx^r_h, vy^r_h)$, human's radius $rad_h$, distance to robot $dr$, angle between robot's velocity vector and human's position vector w.r.t robot $theta\_hr\_rv$, angle between relative goal vector and human's position vector w.r.t robot $theta\_hr\_gr$, awareness $\alpha$ ($\alpha=1.0$ if human is aware and $\alpha=0.0$ if unaware). Let their be N humans in the scene, then $H = \{h_1, h_2, ..., h_N\}$ is the state of all humans in the scene.

\emph{A simplistic model for human's un/awareness towards robot}:
Estimating human's state of awareness towards the robot is assumed to be known, and is out of the scope of this work. However, we briefly discuss how this state can be modelled. 

Consider $\Omega$ to be the direction of gaze/head of the human and $\Gamma$ be the set of simple activities, e.g., $\Psi$ = \{\emph{using cellphone}, \emph{texting}, \emph{reading}, \emph{browsing the cellphone},  \emph{picking up an object}, \emph{...}\}, i.e., $\Gamma \in \Psi$. Then at any time instance $t$, the following heuristics can be used to model a human's ($h$) state of awareness $\alpha_t^h$ towards the robot ($r$): 



\begin{equation}
    \alpha_t^h(\Omega^h, \Gamma^h) = 
    \begin{cases}
      \text{aware-of-robot,} & \text{if } \Gamma^h \notin \Psi  \\
      & \text{   } \cap \text{ FoV}(\Omega^h, \mathbf{p}_{r})\\
      \text{unaware-of-robot,} & \text{if } \Gamma^h \in \Psi  \\
    \end{cases}
\end{equation}

where $\bold{\mathbf{p}_{r}}$ is the robot's pose, FoV($\cdot$) is a function that returns $True$ if the robot ($r$) is in the field-of-view of the human $h$.

\emph{Robot-Human attention}: We have an attention block (Robot-Human attention) which learns to generate importance scores for each human based on robot and humans' states. This attention function is inspired by the scaled dot-product attention in~\cite{transformers}, which computes attention score using a query $Q$ and a key $K$, and applies the normalized score to a value $V$ ( shown by $\bigotimes$ in Figure \ref{fig:2}).

\begin{equation}
    Attention(Q,K,V) = softmax \left(\frac{QK^T}{\sqrt{d_k}}\right) V
\end{equation}
Where $Q = MLP_{q}(H) \in \mathbb{R}^{N \times d_q}$, $K = MLP_{k}(R) \in \mathbb{R}^{1 \times d_k}$, $V = MLP_{v}(V) \in \mathbb{R}^{N \times d_v}$, and $d_q$, $d_k$ and $d_v$ are the dimension of query, key and values, respectively.

\begin{figure}
\center
  \includegraphics[width=0.8\columnwidth]{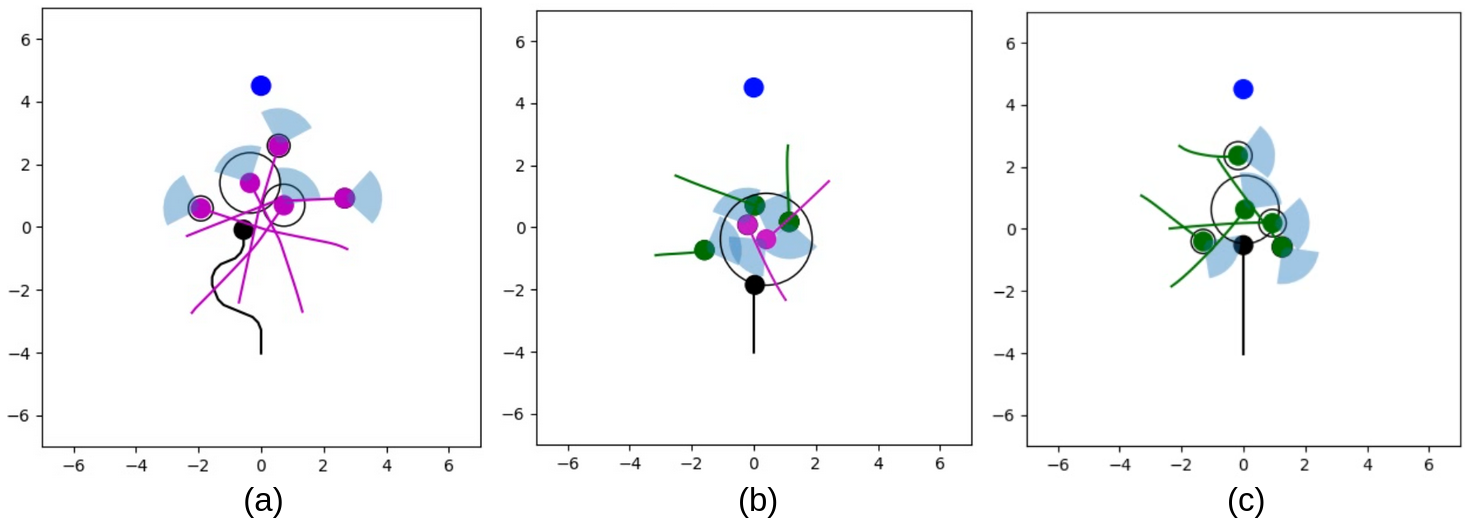}
  \caption{Our attention mechanism operates differently in "Humans unaware" and "Humans aware" cases, as shown in three boxes. The radius of the circle is the proxy for level of attention. In (a), it focuses on the humans the robot is approaching. In (b), it ignores an aware human approaching from the left, assuming they will yield to the robot, and focuses on unaware humans. In (c) where all humans are aware, it prioritizes the closest human. This indicates that the attention mechanism does not trivially learn the awareness state of the humans}
  \label{fig:6}
  \vspace{-0.5cm}
\end{figure}

Figure~\ref{fig:6} shows how the robot attends differently to humans, indicated by the black circle whose radius is a proxy for level of attention.

\emph{Overall state $S_t$ to the state encoder}: After attention  is computed and the score is accordingly applied to human features, we form an overall state $S_t$ which is the concatenation of weighted average of all the human features $Attn\_H = \sum V$, the robot features after being encoded $R\_enc = MLP_{r}(R)$ and the latent distribution parameters $Z_\mu$ and $Z_\sigma$, concatenated and encoded to give $Z_{state} = MLP_{zstate}(Z_\mu \oplus Z_\sigma)$.
\begin{equation}
    S_t = [ Attn\_H \oplus R_{enc} \oplus Z_{state}]
\end{equation}
The overall state $S_t$ is fed to the state encoder which is an MLP. The outcome of the state encoder is then mapped to value $V_t$ and a continuous policy $\pi(a_t|st)$ where the action $a_t$ is the control variable (in this case $\epsilon_t$) obtained as $\mathcal{N}(\epsilon_t |  \epsilon_{ \mu(s_{t}) },  \epsilon_{ \sigma(s_{t}) })$ 
which is used to generate a latent code $Z_{control}$. This latent code is then fed to the decoder $\mathcal{D}$ of the trajectory forecaster to produce next objective-oriented motion vector $\Delta x_{control}$, $\Delta y_{control}$. The motion vector is converted to velocity by dividing by the time step and mapped to discrete action space.
\begin{equation}
    Z_{control} = Z_\mu + Z_\sigma * \epsilon_t 
    \label{eq:zcontrol}
\end{equation}
\begin{equation}
    (\Delta x_{control}, \Delta y_{control})  = \mathcal{D}(Z_{control} * 4.0 )
\end{equation}
\begin{equation}
    (vx, vy)  = \emph{ToDiscreteAction}(\frac{\Delta x_{control}}{\Delta t}, \frac{\Delta y_{control}}{\Delta t})
\end{equation}

Where, $\mathcal{D}$ resembles the decoder of the trajectory forecasting block (red dashed box); note that the entire trajectory forecaster is frozen during training and testing. 

\emph{Architecture details.} The $MLP_q$ and $MLP_v$ are made up of two full-connected (FC) layers whose input and output feature dimensions are 9 (length of human state vector) and 256, respectively with ReLU activation between the two -- [FC(9,256), ReLU, FC(256,256)]. $MLP_k$ and $MLP_r$ are [FC(7,256), ReLU, FC(256,256)], where 7 is length of robot's state vector. The $MLP_{zstate}$ is made as [FC(64,256), ReLU, FC(256,256)] where 64 is the length of the $Z_{state}$ (as length of $Z_\mu$ and $Z_\sigma$ are 32 element vectors). The state encoder is composed as [FC(768, 512), ReLU, FC(512, 512), ReLU, FC(512, 256)]. Finally, the output of the state encoder is mapped to $V(s_t)$ using FC(256,1) and mapped to  $\epsilon_{\mu(s_t)}$ and $\epsilon_{\sigma(s_t)}$ using [FC(256, 32), Tanh]. As we have Tanh activation in the policy whose output range is [-1, 1], we scale the latent code $Z_{control}$ by 4.0 (in Eq. \ref{eq:zcontrol}) to make sure that the almost the entire standard normal distribution is covered, as $\epsilon_t \in \mathcal{N}(0, I)$.

\emph{Reward function}: Our model-free RL training scheme uses a dense reward function that rewards the robot for reaching or approaching the goal and penalizes collisions or getting too close to humans at any time $t$.

\begin{equation}
\label{eq:reward_fn}
  r_t(s_t, a_t) =
    \begin{cases}
      10,  &  \text{if } d_g^t \leq \rho_{robot}\\ 
      -20, &  \text{if } d^t_{min} < 0 \\ 
      4.0(d^t_{min} - 0.25), & \text{if } 0 < d^t_{min} < 0.25 \\
      2.0(d^{t-1}_{goal} - d^{t}_{goal}), & \text{otherwise.}\\
    \end{cases}       
\end{equation}

In the above reward function (Eq. \ref{eq:reward_fn}), $r_t(s_t, a_t)$, at any time instance $t$, $d^t_{min}$ is the distance between the robot and the closest human, $d^t_{goal}$ is the $L2$ norm between the robot and its goal, $\rho_{robot}$ is the robot's radius. 
The value of a state at time $t$, is estimated using the \emph{n-step} return scheme: $V_t(s_t) = \sum_{i=t}^{t+n} ( r_t(s_t, a_t) \cdot \gamma^{i-t}) + \widehat{V_{t+n+1}}$. Here $\gamma$ is the discount factor and  $\widehat{V_{t+n+1}}$ is the network predicted value for the next state.

\emph{Deep reinforcement learning.}
We rely on policy-based model-free reinforcement learning (RL) method, that directly parameterizes the policy $\pi(a|s, \theta)$. We use advantage actor-critic (A2C) framework for optimizing our controller. 

\subsection{Training}

During end-to-end training, \emph{n-steps} trajectories are rolled out from 24 synchronized environments and recorded as a batch for policy updates. The number of aware humans in each environment is randomly chosen. Policy updates employ 5 steps per environment ($n$=5), with discount factor $\gamma$ = $0.90$, and coefficients $\eta =$ 1.0 for policy loss, $\psi =$ 0.001 for entropy term, and $\zeta = $ 0.25 for critic loss. The network is trained for approximately $2\times10^{5}$ episodes, with each episode lasting a maximum of 30$s$ or 120 steps ($\Delta t$ = 0.25$s$).

\begin{table}[h]
\caption{Overall Efficacy}
\label{overall}
\begin{center}
\scalebox{0.71}{
\begin{tabular}{c c c c c c c}
\hline
\textbf{Method} & \textbf{Success(\%)$\uparrow$} & \textbf{NT(s)$\downarrow$} & \textbf{Jerk(m/s$^{3}$)$\downarrow$} & \textbf{TL(m)$\downarrow$} & \textbf{Disc$\downarrow$} & 
\textbf{Soc(m)$\uparrow$}\\
\hline
CRI-Linear~\cite{cri} & 96.8\%  & 12.25 & 0.58 & \textbf{9.917} & 2.53 & 1.34\\

DSRNN~\cite{dsrnn} & 94\%  & 14.51 & 0.17 & 13.48 & 0.46 & 0.62\\

RGL~\cite{rgl} & 97\% & \textbf{12.20} & 0.31 & 10.017& 4.35 & 0.96\\

CrowdNav++~\cite{intentaware} & 82\% & 13.07 &  0.17& 10.95& 0.39 & 1.00\\

DSRNN - D~\cite{dsrnn} & 94.8\% & 14.72 &  0.35& 13.55& 0.58 & 0.60\\

CrowdNav++ - D~\cite{intentaware} & 95\% &  13.60 &  0.43& 12.93& 0.49 & \textbf{1.54}\\
\hline

Ours-WFC-D  & \textbf{98.2}\% & 13.00 & \textbf{0.15} & 10.69&  \textbf{0.37} & 1.01\\
\hline
\end{tabular}}
\end{center}
\end{table}

\begin{table}[h]
\caption{Performance with continuous actions}
\label{continuous}
\begin{center}
\scalebox{0.7}{
\begin{tabular}{c c c c c c c}
\hline
\textbf{Method} & \textbf{Success(\%)$\uparrow$} & \textbf{NT(s)$\downarrow$} & \textbf{Jerk(m/s$^{3}$)$\downarrow$} & \textbf{TL(m)$\downarrow$} & 
\textbf{Soc(m)$\uparrow$} & \textbf{Disc$\downarrow$}
\\
\hline
Ours-WFC-C & \textbf{98.4\%}  & \textbf{11.55} & \textbf{0.13} & \textbf{10.80} & \textbf{1.16} & \textbf{0.29}\\

CrowdNav++~\cite{intentaware} & 82\% & 13.07 &  0.17& 10.95& 1.00 & 0.39 \\

DSRNN~\cite{dsrnn} & 94\%  & 14.51 & 0.17 & 13.48 &  0.62 & 0.46 \\
\hline
\end{tabular}}
\end{center}
\end{table}

\section{EXPERIMENTS}
\subsection{Simulation environment}
We utilize the CRI-Linear\footnote{https://github.com/vita-epfl/CrowdNav}~\cite{cri} simulation environment, controlling agents (considered as humans) with ORCA~\cite{orca}. To ensure fair comparisons, we adopt the circle crossing scenario, featuring N = 5 humans randomly placed on a 4.5m radius circle. Unlike the conventional discomfort region (0.25m radius) around humans, we randomly perturb their starting positions with the goal in the opposite direction. Agents have a maximum speed of 1 $m/s$, and the simulation runs at 4 steps/sec ($\Delta t=0.25 sec$). We evaluate our method using the 500 random test cases from~\cite{cri} simulation framework. To examine scenarios with aware and unaware humans, we augment human agents with the awareness attribute, allowing us to control their reciprocation to the robot's motion. 


\subsection{Evaluation Metrics}

We evaluate our method on the following metrics:
\subsubsection{Success}
The fraction of episodes in \% where the robot reached its goal to the total number of episodes.
\subsubsection{Average Navigation Time(NT)}
The average time taken by the robot to reach its goal from the starting position across all episodes
\subsubsection{Average Jerk} The average change in acceleration per unit time across all episodes. Lower values indicate smoother movements.
\subsubsection{Average Trajectory Length(TL)} The average distance traveled by the robot from the starting position to the goal across all successful episodes.
\subsubsection{Average Discomfort(Disc)} The ratio of the number of timesteps where the robot enters the discomfort region(0.25m) of at least one human to the total number of timesteps.
\subsubsection{Sociability(Soc)} 
The average closest distance between a human and the robot across all episodes in which the robot is in navigable field of view of the human. Not going too close in the humans FOV indicates robot's ability to anticipate human motion. A sociability value $\geq$1.0 is preferable as based on theory of proxemics \cite{proxemics_survey} which states that an area with 1m radius circle around human is approximately equal to the personal space region.
\subsubsection{$<$28$^{\circ}$} 
This refers to the proportion of successive steps(in \%) in which the angle deviation is  $<$28$^{\circ}$.
\subsubsection{Mean/Std$^{\circ}$} 
This pertains to the mean and the standard deviation of the angular difference between successive steps. Lower values indicate legible trajectories.

\begin{figure*}
\center
  \includegraphics[width=0.7\textwidth]{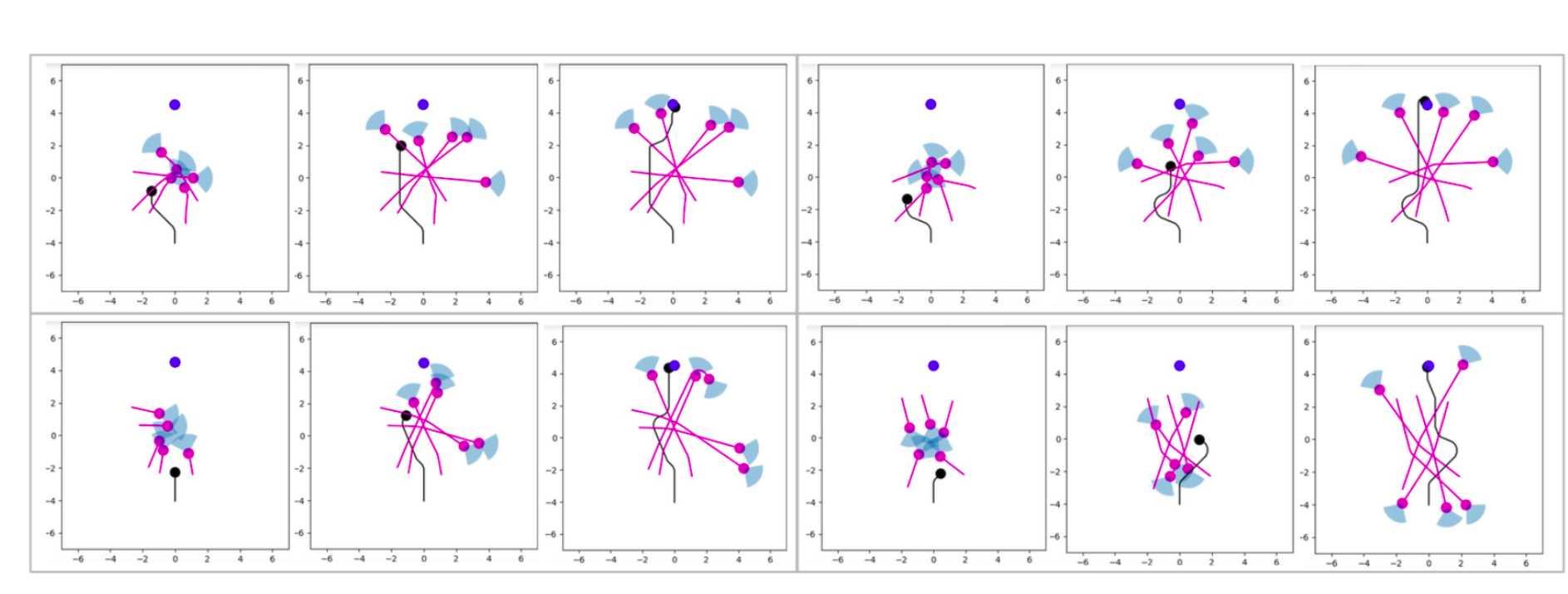}
  \caption{Our method's (Ours-WFC-D) efficacy is demonstrated through a visual representation of robot trajectories across four episodes (two per row). The robot, unaware human and the goal is shown by black, purple and blue discs respectively.We show snapshot of robot's trajectory for three different time steps. It can be seen that our policy leads to smooth, legible and anticipatory robot motion. We  attributed the performance to our innovative idea of generating of actions through the use of social motion latent space.}
  \label{fig:mrd}
\end{figure*}
\begin{figure*}
\center
  \includegraphics[width=0.8\textwidth]{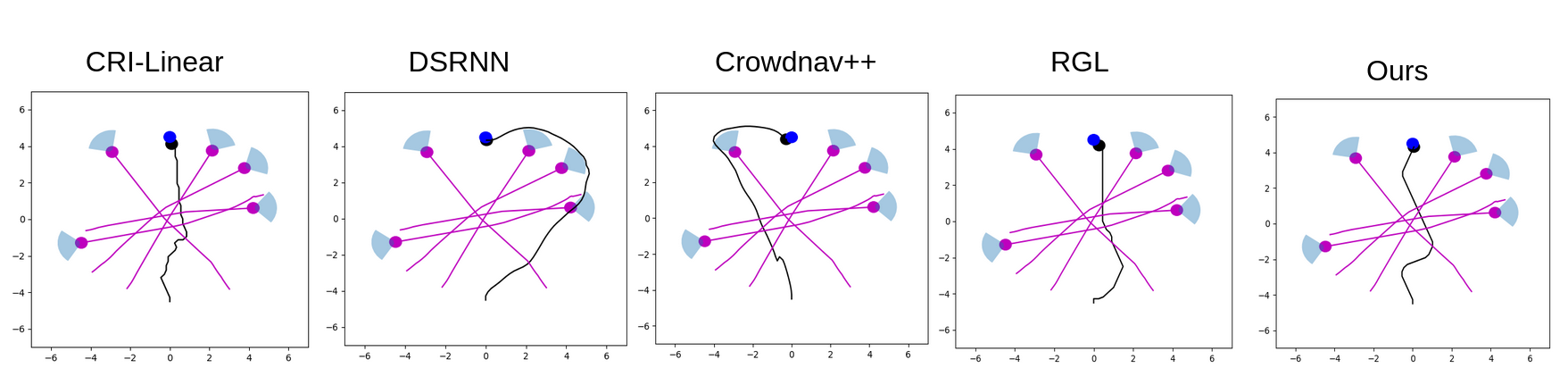}
  \caption{Our approach (Ours-WFC-D) is compared with several baselines. Our policy demonstrates the ability to balance between  length(and time) and legibility(smoother/less jerk) of the trajectories. CRI-Linear and RGL produce jerky movements while achieving short trajectories, and CrowdNav++ and DSRNN produce longer trajectories while achieving relatively smoother ones. Our approach strikes a balance, resulting in efficient and smooth trajectories.}
  \label{fig:4}
\end{figure*}
\begin{figure}

  \includegraphics[width=\columnwidth]{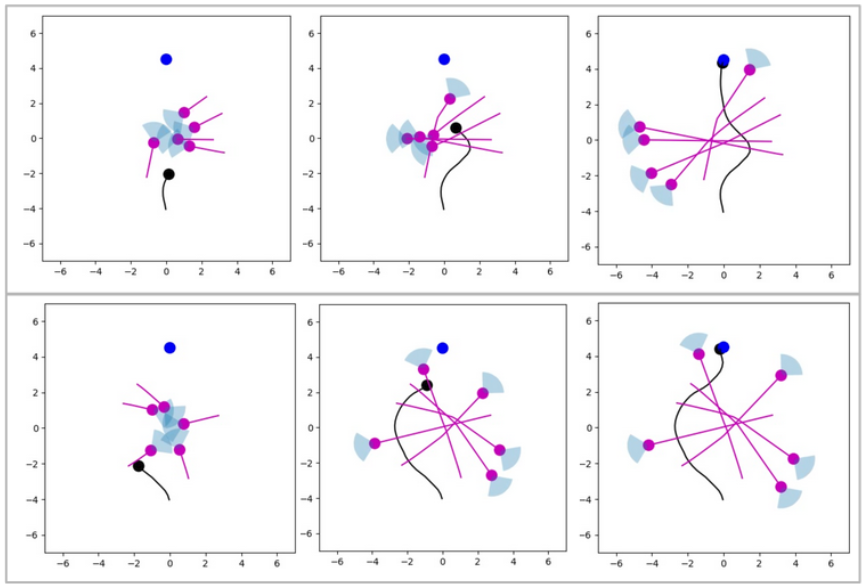}
  \caption{As we are generating actions from social motion latent space which is continuous in nature, the continuous version of our method (Ours-WFC-C) works equally well with continuous actions. We show snapshot of two different episodes which makes it clear we get smooth and anticipatory trajectory despite the model not being trained with discrete action-space.}
  \label{fig:5}

\end{figure}

\begin{figure}
\vspace{0.5cm}
\center
  \includegraphics[width=0.8\columnwidth]{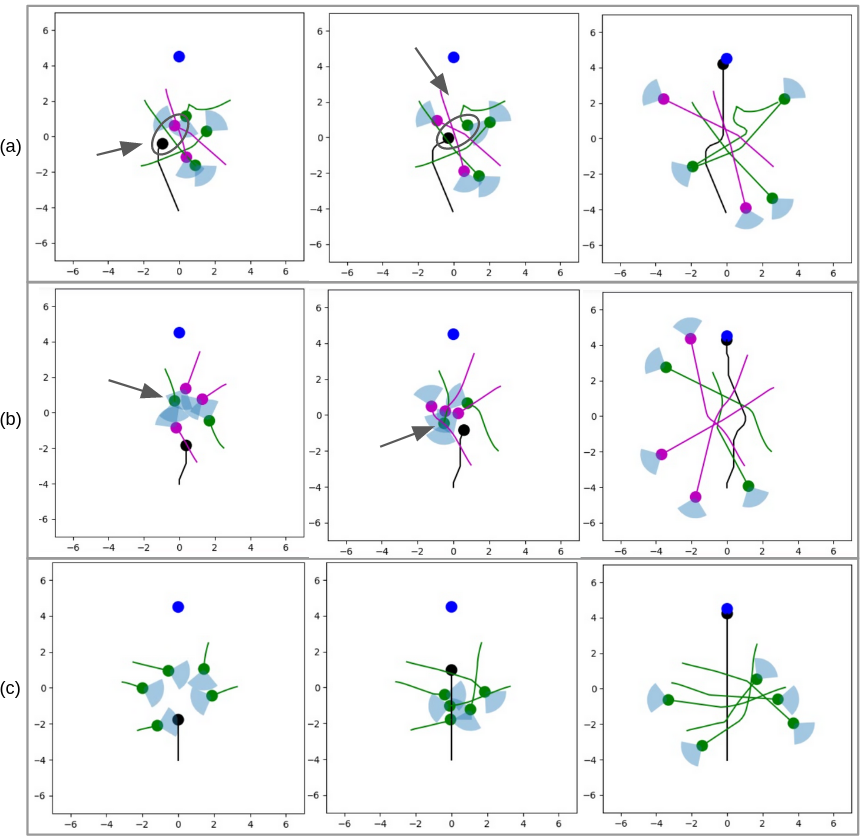}
  \caption{The figure illustrates how human awareness impacts social navigation, with green representing aware humans and purple representing unaware humans. Each row shows a different situation with varying numbers of aware humans. In the first scenario, the robot deviates from the path of the unaware human, but an aware human makes way for the robot in the second scenario. The second scenario depicts an aware human deviating to let the robot pass, and in the last scenario, all humans are aware and make way for the robot to take a straight path towards the goal.}
  \label{fig:7}
\end{figure}
\subsection{Experimentation Results}
In this particular section, we will be showcasing our findings in both quantitative and qualitative forms.

\subsubsection{Baselines}
 Our approach will be compared with baslines, including DSRNN~\cite{dsrnn}, CRI-Linear~\cite{cri}, RGL~\cite{rgl}, and CrowdNav++(has future collision penalty in its reward structure)~\cite{intentaware}, using the metrics mentioned earlier. The method CRI~\cite{cri} assumes that the next step ground truth state is provided
during training which is an unrealistic assumption. We use a more realistic version of this, CRI-Linear~\cite{cri} (like done in~\cite{rgl}) which utilizes a
linear motion model where agents maintain their velocities
as in their previous state. CRI-Linear~\cite{cri} and RGL~\cite{rgl} uses discrete action spaces. DSRNN~\cite{dsrnn} and Crowdnav++~\cite{intentaware} uses continuous actions by default. To show the results of the discrete versions of versions of DSRNN and Crowdnav++ we map their continuous actions to the closest action in the discrete action space these are mentioned as DSRNN - D and Crowdnav++ - D.

\begin{table}[h]
\caption{Smoothness Statistics}
\label{smoothness}
\begin{center}
\scalebox{0.80}{
\begin{tabular}{c c c c c}
\hline
\textbf{Method}  & \textbf{$<$28$^{\circ}$$\uparrow$} & \textbf{Mean/Std$^{\circ}$$\downarrow$}
& \textbf{NT(s)$\downarrow$}
& \textbf{TL(m)$\downarrow$}\\
\hline
CRI-Linear~\cite{cri} & 80.3\%  & 24.57/41.80 & 12.25 & \textbf{9.917} \\

RGL~\cite{rgl} & 92.00\%  & 14.31/35.92  & 12.20 & 10.02\\

DSRNN~\cite{dsrnn} & 89.80\%  & 20.83/56.34 & 14.51 & 13.48 \\

CrowdNav++~\cite{intentaware} & 87.90\%  & 20.82/59.35  & 13.07 & 10.95 \\

DSRNN - D~\cite{dsrnn} & 89.55\%  & 22.15/54.51 & 14.72 & 13.55 \\

CrowdNav++ - D~\cite{intentaware}& 79.46\%  & 32.78/62.15  & 13.60 & 12.93 \\
\hline
Ours - WFC-C  & \textbf{96.24\%}  & 9.02/\textbf{25.19} & \textbf{11.55} & 10.80  \\

Ours - WFC-D  & 95.99\%  & \textbf{8.75}/27.98 & 13.00 & 10.69 \\

\hline
\end{tabular}}
\end{center}
\end{table}

\begin{table*}[h]
\vspace*{10px}
\label{speed}
\begin{center}
\scalebox{0.8}{
\begin{tabular}{c c c c c c c}
\hline
\textbf{Method (max speed)} & \textbf{Success(\%)$\uparrow$} & \textbf{NT(s)$\downarrow$} & \textbf{Jerk(m/s$^{3}$)$\downarrow$} & \textbf{TL(m)$\downarrow$} & 
\textbf{Soc(m)$\uparrow$} & \textbf{Disc$\downarrow$}\\
\hline
Ours-WFC-C
(1m/s)
 & \textbf{98.4\%}  & \textbf{11.55} & \textbf{0.13} & \textbf{10.80} & \textbf{1.16}
 & \textbf{0.29}\\

CrowdNav++ (1m/s)~\cite{intentaware} & 82\%  & 13.07 & 0.17 & 10.95
 & 1.00 & 0.39\\

DSRNN (1m/s)~\cite{dsrnn} & 94\% & 14.51 & 0.17 & 13.48 & 0.62 & 0.46\\
\hline
Ours-WFC-C
(0.75m/s)
 & \textbf{98.8\%} & \textbf{14.09}
 &  \textbf{0.07} & 10.27 & 1.30 & 0.21\\

CrowdNav++ (0.75m/s) (0.5 goal radius)*\footnote{* indicates we increase the goal reaching radius distance to be fair with baselines.}~\cite{intentaware}& 96\% & 15.23 &  0.09 & ~\textbf{10.12} & \textbf{1.44} & \textbf{0.17}\\

DSRNN (0.75m/s) (0.5 goal radius)* ~\cite{dsrnn}& 91\% &  18.68 &  0.07 & 12.70 & 0.75 & 1.17\\
\hline

Ours-WFC-C (0.5m/s)
  & \textbf{96.4\%} & \textbf{19.50} & 0.06 & 9.83 &  \textbf{1.50} & \textbf{0.39}\\
CrowdNav++ (0.5m/s) (0.5 goal radius)*~\cite{intentaware}
  & 93\% & 20.18 & 0.03 & \textbf{9.23} &  1.41 & 0.42\\
  DSRNN (0.5m/s) (1.0 goal radius)*~\cite{dsrnn}
  & 76\% & 20.41 & \textbf{0.02} & 9.54 &  1.21 & 1.91\\
  
\hline
\end{tabular}}
\caption{Speed constraint statistics}
\end{center}
\end{table*}

\begin{table}[h]
\caption{Ablation}
\label{ablation}
\begin{center}
\scalebox{0.84}{
\begin{tabular}{c c c c c}
\hline
\textbf{Method} & \textbf{Success(\%)$\uparrow$} &  \textbf{Jerk(m/s$^{3}$)$\downarrow$} & 
\textbf{$<$28$^{\circ}$$\uparrow$} & \textbf{Mean/Std$^{\circ}$$\downarrow$} \\
\hline
Ours-WFC-D
 &  98.2\%  & \textbf{0.152} & \textbf{95.99} & \textbf{8.75/27.98}\\

Ours-Abl-D &  98.8\% &  0.178  & 94.97\% & 9.51/28.34\\

Ours-NoSVAE-D &  \textbf{99.4\%} &  0.606 & 76.41\% & 34.35/72.70 \\
\hline
\end{tabular}}
\end{center}
\end{table}

\begin{table*}[h]
\vspace*{10px}
\label{awarenesstable}
\begin{center}
\scalebox{0.80}{
\begin{tabular}{c c c c c c c c c c} 
\hline
\textbf{Method} & \textbf{Awareness} & \textbf{Success(\%)$\uparrow$} & \textbf{NT(s)$\downarrow$} & \textbf{Jerk(m/s$^{3}$)$\downarrow$} & \textbf{TL(m)$\downarrow$} & \textbf{Soc(m)$\uparrow$} & 
\textbf{$<$28$^{\circ}$$\uparrow$} & \textbf{Mean/Std$^{\circ}$$\downarrow$} & \textbf{Disc$\downarrow$} \\
\hline
 Ours-WFC-D & 0\% & 98.2\% & 13.0 & 0.152 & 10.69 & \textbf{1.01} & 95.99\% & 8.75/27.98 & 0.37 \\

 Ours-WFC-D & 60\% & 98.8\% & 11.77 & 0.124 & 9.79 & \textbf{1.01} & 97.55\% & 5.94/20.44 & 0.27 \\

 Ours-WFC-D & 100\% & \textbf{100\%} & \textbf{10.38} & \textbf{0.007} & \textbf{8.80} & 0.93 & \textbf{99.04\%} & \textbf{0.22/2.30} & \textbf{0.00} \\
 \hline
 Ours-WFC-C & 0\% & 98.4\% & 11.55 & 0.13 & 10.80 & \textbf{1.16} & 96.24\% & 9.02/25.19 & 0.29 \\

Ours-WFC-C & 60\% & 98.4\% & 10.77 & 0.10 & 9.93 & 1.00 & \textbf{98.20\%} & 6.41/15.86 & 0.33 \\

 Ours-WFC-C & 100\% & \textbf{100\%} & \textbf{10.43} & \textbf{0.05} & \textbf{8.92} & 0.84 & 87.89\% & \textbf{2.48/2.99} & \textbf{0.01} \\

 \hline
 Ours-WFC & 100\% (perceived unaware) & \textbf{100\%} & 12.15 & 0.12 & \textbf{10.18} & \textbf{1.06} & 96.67\% & \textbf{6.93}/26.67 & \textbf{0.00} \\

 Ours-WFC-C & 100\% (perceived unaware) & \textbf{100\%} & \textbf{11.13} & \textbf{0.11} & 10.44 & 0.96 & \textbf{97.10}\% & 7.58/\textbf{21.87} & 0.01 \\
\hline

\end{tabular}}
\caption{EFFECT OF INCORPORATING HUMAN AWARENESS INTO THE SOCIAL NAVIGATION FRAMEWORK}
\end{center}
\end{table*}

\begin{table*}
\caption{Negligible degradation due to absence of future collision check reward}
\label{degredation}
\begin{center}
\scalebox{0.8}{
\begin{tabular}{c c c c c c c c c}
\hline
\textbf{Method} & \textbf{Success(\%)$\uparrow$} & \textbf{NT(s)$\downarrow$} & \textbf{Jerk(m/s$^{3}$)$\downarrow$} & \textbf{TL(m)$\downarrow$} & \textbf{Soc(m)$\uparrow$} & 
\textbf{$<$28$^{\circ}$$\uparrow$} & \textbf{Mean/Std$^{\circ}$$\downarrow$} &
\textbf{Disc$\downarrow$}\\
\hline
Ours-WFC-D
 &  98.2\% & 13.0 & 0.15 & 10.69 & 1.01 & 95.99 & 8.75/27.98 & 0.37\\
Ours-Abl-D
 &  \textbf{98.8\%} & 12.77 & 0.17 & \textbf{10.51} & 1.05 & 94.97 & 9.51/28.34 & 0.34\\
CrowdNav++ - D~\cite{intentaware}
 &  95\% & 13.60 & 0.43 & 12.93 & \textbf{1.54} & 79.46 & 32.78/62.15 & 0.49\\
Ours-WFC - C
 &  98.4\% & \textbf{11.55} & \textbf{0.13} & 10.80 & 1.16 & \textbf{96.24} & 9.02/\textbf{25.19} & 0.29\\
Ours-Abl - C
 &  98.6\% & 11.64 & 0.18 & 10.66 & 1.17 & 93.55 & 11.82/30.61 & \textbf{0.23}\\
CrowdNav++ - C~\cite{intentaware}
 &  82\% & 13.07 & 0.17 & 10.95 & 1.00
 & 87.90 & \textbf{0.82}/59.35 & 0.39\\
 
\hline
\end{tabular}}
\end{center}
\end{table*}

\subsubsection{Overall Efficacy}
The purpose of our analysis is to determine the effectiveness of our method. Our results, as outlined in Table \ref{overall}, indicate that our method outperforms the other approaches across several metrics. Specifically, our method achieved the best results for success rate, average jerk, and average discomfort.

Our method(Ours-WFC-D) achieved a 98.2\% success rate, indicating it is safe and has strong goal-reaching ability. Our approach resulted in an average jerk of 0.15m/s$^{3}$, indicating legible trajectories compared to other approaches. Our analysis also showed a low discomfort frequency to other humans, with an average of 0.37, the best among the compared approaches.

Furthermore, despite its safe nature, our robot exhibited comparable navigation time and trajectory length to CRI-Linear~\cite{cri} and RGL~\cite{rgl}. These methods exhibit jerky trajectories that are not particularly legible. We also noted that our approach had the second-best sociability score of 1.01m, only losing out to CRI-Linear~\cite{cri}, which has a score of 1.34m. This indicates that our method is able to maintain a safe distance from other agents despite not having any reward for this property.

In Fig~\ref{fig:mrd}, we present the qualitative results of our method. It is evident from the figure that our approach has the capability of producing trajectories that are both socially acceptable and smooth. It can be seen how the robot anticipate human's future motion and accordingly change its trajectory to avoid collision. 

In Fig~\ref{fig:4}, we show a qualitative comparison of our method with various baselines. Our results reveal that our approach generates trajectories that are not only smoother than the baselines, but also short in length. This shows that our method is able to strike balance between smooth and short trajectories.
\\

\subsubsection{Exploiting social motion latent space of a forecasting network}
In this section, we show how generating actions from the social latent space of a motion forecasting network has advantages. Quality of robot trajectories depends on properties such as smoothness, operating in discrete/continuous action spaces, and performing well with limited speed. We compare our approach to baselines on these properties and find that our method outperforms them quantitatively and qualitatively, demonstrating the effectiveness of generating actions from the social motion latent space. 

Navigating using continuous actions is important as it allow for smooth movement through the environment, enabling precise adjustments to be made in response to environmental changes.  Our method, called Ours-WFC-C, demonstrates the ability to navigate using continuous actions despite being trained using a discrete action space. In Table \ref{continuous} we compare our approach to baselines, including Crowdnav++~\cite{intentaware} and DSRNN~\cite{dsrnn}, and observe that our method outperforms them on all six evaluation metrics. Specifically, our method achieves a 98.4\% success rate, indicating a high capability to reach the goal. The trajectories obtained by our robot are efficient, with a low navigation time of 11.55s and shorter trajectory lengths of 10.80m. Our method achieves a low jerk of 0.13m/s$^{3}$, demonstrating the ability to produce smooth trajectories compared to the baselines. Furthermore, our method achieves trajectories that are social in nature, with a sociability of 1.16m and an average discomfort frequency of 0.29. Overall, our results demonstrate that the use of the social latent space of a motion forecasting network, combined with the ability to navigate using both continuous and discrete action spaces, enhances the performance of our method compared to baselines.
Fig~\ref{fig:5} of our study showcases the qualitative outcomes of two different episodes when the robot takes continuous actions.

Smooth robot trajectories are vital for efficient navigation, offering benefits such as energy efficiency, safety, precision, robustness, and human-like behavior.  In Table \ref{smoothness} we compare the smoothness statistics of our method with baselines on four performance metrics, including the fraction of robot steps with angle deviation $<$28$^{\circ}$, Mean/Std$^{\circ}$ of angle change between time steps, navigation time, and average trajectory length. Our two variants, Ours-WFC-C and Ours-WFC-D, operate using continuous and discrete actions, respectively. Our results indicate that our method produces smoother trajectories, achieving 96.24\% and 95.99\% for the first metric $<$28$^{\circ}$. Furthermore, Ours-WFC-D has the best mean deviation of 8.75$^{\circ}$, while Ours-WFC-C has the best standard deviation of 25.19$^{\circ}$. Ours-WFC-C also achieved faster navigation time of 11.55s compared to baselines. 

Socially appropriate robot navigation, even at lower speeds, is crucial for navigating complex social scenarios involving humans and other robots. Lower speeds decrease the risk of accidents and collisions, resulting in safer and more manageable navigation. Moreover, robot's could also enforce speed constraints that the policy should be able to handle. In Table IV, we compared the performance of our method with two baselines, DSRNN~\cite{dsrnn} and CrowdNav++~\cite{intentaware}, while testing three different maximum speeds: 1m/s, 0.75m/s, and 0.5m/s. Our method demonstrated superiority over both baselines on all metrics at the maximum speed of 1m/s. When the maximum speed was restricted to 0.75m/s, our method outperformed DSRNN~\cite{dsrnn} on all metrics and surpassed CrowdNav++~\cite{intentaware} on three metrics. At a maximum speed of 0.5m/s, our method outperformed both baselines on four metrics, producing trajectories that were more efficient, safer, and socially adept. 

For cases where maximum speed is capped to 0.75m/s and 0.5m/s we add a * next to CrowdNav++ and DSRNN. This indicates that we have increased the goal radius to 0.5m and 1m for these methods for fairness. Without this, both these methods resulted in the robot timing out very close to the goal for many episodes(30\% - 40\%) when the goal radius was set to the original value of 0.3m.

We would like to emphasize that despite not incorporating future collision reward in our method (Ours-WFC-D), we are able to achieve anticipatory robot motion as we generate action from a social motion latent space learned by a social trajectory forecaster, which inherently encodes this information. To validate this, we train a version of our model with future collision check in the reward structure (Ours-Abl-D) similar to Crowdnav++~\cite{intentaware}. Table~\ref{degredation} shows that our method Ours-WFC-D performs comparable on all parameters to Ours-Abl-D and outperforms Crowdnav++~\cite{intentaware} (both discrete and continuous). The performance is consistent even for the continuous versions (Ours-WFC-C and Ours-Abl-C). 


\subsubsection{Ablation}
To understand the benefit of generating robot actions from a social motion latent space, we trained a variant of our model that directly produces probabilites for the used discrete actions -- Ours-NoSVAE-D. From the ablation Table \ref{ablation} it is clear that with negligble degradation in success rate, our model Ours-WFC-D outperforms both Ours-Abl-D and Ours-NoSVAE-D. Our model (Ours-WFC-D) achieves smooth trajectories even without a future collision reward, as used in Crowdnav++~\cite{intentaware}; avoiding future collision check would prevent our policy from getting effected by the prediction errors of the trajectory forecaster.

\subsubsection{Social Awareness}
In Table VI we demonstrate the benefits and outcomes of integrating human awareness into the robot state. Three sets of results are shown, where  in the first two sets we show the effect of increasing the number of humans being aware. In the third set we show the effect of not incorporating awareness in the robot state.

In the first set, we demonstrate the results of our method using Discrete action space Ours-WFC-D. It is noteworthy that an increase in the number of humans' awareness leads to an overall improvement in most metrics. When all humans are aware (100\%), our method achieves a success rate of 100\%. Moreover, the robot's trajectories are improved, with a navigation time of 10.38s and a trajectory length of 8.80m. Since humans tend to yield to the robot, the robot needed to take fewer deviations, resulting in less jerk of 0.007m/s$^{3}$ and minimal discomfort. In addition, we observe low angular deviations as shown in columns 8($<$28$^{\circ}$) and 9(Mean/Std$^{\circ}$).


Lastly, in the third set where all humans are aware, we demonstrate the performance of both Ours-WFC-D and Ours-WFC-C when all human are aware but the robot perceives them all unaware. In such a case, important feature to observe is that while success rate is not affected, the navigation time increases and robot keeps larger distance to humans. This shows that it is needed for the robot to be cognizant of the humans' awareness (i.e. have it as a state element) to benefit from their positive interactions i.e. yielding to the robot.

Fig~\ref{fig:7} displays qualitative results from different episodes, with each row representing a distinct scenario. The first row pertains to the case where three humans exhibit social awareness, the second row showcases two aware humans, and the final row highlights the instance where all humans demonstrate awareness. Within the first row, the unaware and aware humans are distinguished by purple and green colors, respectively, in the first and second box. It is observed that the unaware human does not yield to the robot in the first box, continuing along its current path. In contrast, the aware human in the second box provides way to the robot, enabling a more efficient trajectory to the goal. Notably, an increase in the number of aware humans results in a corresponding improvement in the quality of the robot's trajectory.

\section{CONCLUSIONS AND FUTURE WORKS}

This paper introduces a novel social robot navigation method utilizing a social motion latent space, resulting in smoother and socially-aware trajectories. Emphasizing the significance of human awareness, it improves navigation speed and success rates. Future research directions involve generating collision-free trajectories from the latent space using contrastive learning and adapting the motion forecasting network during deployment for newer scenarios~\cite{hansen2021deployment}.

\bibliographystyle{IEEEtran}
\bibliography{IEEEexample}

\begin{thebibliography}{10}
\providecommand{\url}[1]{#1}
\csname url@rmstyle\endcsname
\providecommand{\newblock}{\relax}
\providecommand{\bibinfo}[2]{#2}
\providecommand\BIBentrySTDinterwordspacing{\spaceskip=0pt\relax}
\providecommand\BIBentryALTinterwordstretchfactor{4}
\providecommand\BIBentryALTinterwordspacing{\spaceskip=\fontdimen2\font plus
\BIBentryALTinterwordstretchfactor\fontdimen3\font minus
  \fontdimen4\font\relax}
\providecommand\BIBforeignlanguage[2]{{%
\expandafter\ifx\csname l@#1\endcsname\relax
\typeout{** WARNING: IEEEtran.bst: No hyphenation pattern has been}%
\typeout{** loaded for the language `#1'. Using the pattern for}%
\typeout{** the default language instead.}%
\else
\language=\csname l@#1\endcsname
\fi
#2}}

\bibitem{orca}
J.~van~den Berg, M.~Lin, and D.~Manocha, ``Reciprocal velocity obstacles for
  real-time multi-agent navigation,'' 05 2008, pp. 1928--1935.

\bibitem{sf}
D.~Helbing and P.~Molnar, ``Social force model for pedestrian dynamics,''
  \emph{Physical Review E}, vol.~51, 05 1998.

\bibitem{dsrnn}
S.~Liu, P.~Chang, W.~Liang, N.~Chakraborty, and K.~Driggs-Campbell,
  ``Decentralized structural-rnn for robot crowd navigation with deep
  reinforcement learning,'' in \emph{IEEE International Conference on Robotics
  and Automation (ICRA)}, 2021, pp. 3517--3524.

\bibitem{rgl}
C.~Chen, S.~Hu, P.~Nikdel, G.~Mori, and M.~Savva, ``Relational graph learning
  for crowd navigation,'' in \emph{IROS}, 2020.

\bibitem{cri}
C.~Chen, Y.~Liu, S.~Kreiss, and A.~Alahi, ``Crowd-robot interaction:
  Crowd-aware robot navigation with attention-based deep reinforcement
  learning,'' in \emph{ICRA}, 2019.

\bibitem{intentaware}
S.~Liu, P.~Chang, Z.~Huang, N.~Chakraborty, K.~Hong, W.~Liang,
  D.~Livingston~McPherson, J.~Geng, and K.~Driggs-Campbell, ``Intention aware
  robot crowd navigation with attention-based interaction graph,'' in
  \emph{IEEE International Conference on Robotics and Automation (ICRA)}, 2023.

\bibitem{samdrl}
Y.~F. Chen, M.~Everett, M.~Liu, and J.~How, ``Socially aware motion planning
  with deep reinforcement learning,'' 09 2017, pp. 1343--1350.

\bibitem{frozone}
A.~Sathyamoorthy, U.~Patel, T.~Guan, and D.~Manocha, ``Frozone: Freezing-free,
  pedestrian-friendly navigation in human crowds,'' \emph{IEEE Robotics and
  Automation Letters}, vol.~PP, pp. 1--1, 05 2020.

\bibitem{mpc}
T.~Badgwell and J.~Qin, \emph{Model-Predictive Control in Practice}, 01 2013,
  pp. 1--6.

\bibitem{bspline}
Y.~Chen, W.~Dong, and Y.~Ding, ``An efficient method for collision-free and
  jerk-constrained trajectory generation with sparse desired way-points for a
  flying robot,'' \emph{Science China Technological Sciences}, vol.~64, 06
  2021.

\bibitem{jb}
R.~Zhao and D.~Sidobre, ``Trajectory smoothing using jerk bounded shortcuts for
  service manipulator robots,'' in \emph{2015 IEEE/RSJ International Conference
  on Intelligent Robots and Systems (IROS)}, 2015, pp. 4929--4934.

\bibitem{0386}
L.~Liu, D.~Dugas, G.~Cesari, R.~Siegwart, and R.~Dube, ``Robot navigation in
  crowded environments using deep reinforcement learning,'' 10 2020, pp.
  5671--5677.

\bibitem{54}
\BIBentryALTinterwordspacing
S.~B. Banisetty and D.~Feil{-}Seifer, ``Towards a unified planner for
  socially-aware navigation,'' \emph{CoRR}, vol. abs/1810.00966, 2018.
  [Online]. Available: \url{http://arxiv.org/abs/1810.00966}
\BIBentrySTDinterwordspacing

\bibitem{zombie}
J.~Wu, Y.~Tamura, Y.~Wang, H.~Woo, A.~Moro, A.~Yamashita, and H.~Asama,
  ``Smartphone zombie detection from lidar point cloud for mobile robot
  safety,'' \emph{IEEE Robotics and Automation Letters}, vol.~5, no.~2, pp.
  2256--2263, 2020.

\bibitem{56}
R.~Meerhoff, J.~Bruneau, A.~Vu, A.-H. Olivier, and J.~Pettre, ``Guided by gaze:
  Prioritization strategy when navigating through a virtual crowd can be
  assessed through gaze activity,'' \emph{Acta Psychologica}, vol. 190, pp.
  248--257, 10 2018.

\bibitem{57}
J.~Croft and D.~Panchuk, ``Watch where you're going? interferer velocity and
  visual behavior predicts avoidance strategy during pedestrian encounters,''
  \emph{Journal of Motor Behavior}, vol.~50, pp. 1--11, 09 2017.

\bibitem{59}
P.~Trautman, J.~Ma, R.~Murray, and A.~Krause, ``Robot navigation in dense human
  crowds: Statistical models and experimental studies of human-robot
  cooperation,'' \emph{The International Journal of Robotics Research},
  vol.~34, 02 2015.

\bibitem{60}
C.~Vassallo, A.-H. Olivier, P.~Souères, A.~Crétual, O.~Stasse, and J.~Pettre,
  ``How do walkers avoid a mobile robot crossing their way?'' \emph{Gait \&
  Posture}, vol.~51, 09 2016.

\bibitem{64}
R.~Paulin, T.~Fraichard, and P.~Reignier, ``Using human attention to address
  human–robot motion,'' \emph{IEEE Robotics and Automation Letters}, vol.~4,
  pp. 2038--2045, 2019.

\bibitem{65}
T.~Fraichard, R.~Paulin, and p.~reignier, ``Human-robot motion: An
  attention-based navigation approach,'' vol. 2014, 08 2014.

\bibitem{socvae}
P.~Xu, J.-B. Hayet, and I.~Karamouzas, ``Socialvae: Human trajectory prediction
  using timewise latents,'' in \emph{Computer Vision -- ECCV 2022}.\hskip 1em
  plus 0.5em minus 0.4em\relax Cham: Springer Nature Switzerland, 2022, pp.
  511--528.

\bibitem{brojoda}
S.~Sinha, B.~Bhowmick, A.~Sinha, and A.~Das, ``System and method for tracking
  body joints,'' Apr.~20 2021, uS Patent 10,980,447.

\bibitem{boch}
E.~Bochinski, T.~Senst, and T.~Sikora, ``Extending iou based multi-object
  tracking by visual information,'' in \emph{2018 15th IEEE International
  Conference on Advanced Video and Signal Based Surveillance (AVSS)}, 2018, pp.
  1--6.

\bibitem{vae}
D.~P. Kingma and M.~Welling, ``{Auto-Encoding Variational Bayes},'' in
  \emph{2nd International Conference on Learning Representations, {ICLR} 2014,
  Banff, AB, Canada, April 14-16, 2014, Conference Track Proceedings}, 2014.

\bibitem{motionvae}
H.~Y. Ling, F.~Zinno, G.~Cheng, and M.~van~de Panne, ``Character controllers
  using motion vaes,'' \emph{ACM Trans. Graph.}, vol.~39, no.~4, 2020.

\bibitem{transformers}
\BIBentryALTinterwordspacing
A.~Vaswani, N.~M. Shazeer, N.~Parmar, J.~Uszkoreit, L.~Jones, A.~N. Gomez,
  L.~Kaiser, and I.~Polosukhin, ``Attention is all you need,'' in \emph{NIPS},
  2017. [Online]. Available:
  \url{https://api.semanticscholar.org/CorpusID:13756489}
\BIBentrySTDinterwordspacing

\bibitem{proxemics_survey}
J.~Rios-Martinez, A.~Spalanzani, and C.~Laugier, ``From proxemics theory to
  socially-aware navigation: A survey,'' \emph{IJSR}, 2015.

\bibitem{hansen2021deployment}
N.~Hansen, R.~Jangir, Y.~Sun, G.~Alenyà, P.~Abbeel, A.~A. Efros, L.~Pinto, and
  X.~Wang, ``Self-supervised policy adaptation during deployment,'' in
  \emph{International Conference on Learning Representations}, 2021.

\end{thebibliography}
\end{document}